\documentclass[runningheads]{llncs}
\usepackage{esvect}
\usepackage[T1]{fontenc}
\usepackage{graphicx}
\usepackage{amsmath}
\usepackage{booktabs} 
\usepackage{multirow} 
\usepackage{array}
\usepackage{comment}
\usepackage[colorlinks=true, allcolors=blue]{hyperref}

\usepackage{tabularx}
\usepackage{cleveref}
\usepackage{adjustbox}
\usepackage{xcolor}
\usepackage{rotating}
\usepackage[super]{nth}
\usepackage{mathtools}
\usepackage{hyphenat}
\usepackage[british]{babel} 
\usepackage{csquotes}
\usepackage{setspace}
\usepackage{pgfplots}
\pgfplotsset{compat=1.18}

\usepackage{tikz}
\usepackage{etoolbox}
\usepackage{adjustbox}
\usepackage{todonotes}
\usepackage{subfig}
\usepackage{amsmath,amssymb,amsfonts}

\authorrunning{Wael Korani et al.}

\begin{document}
\title{Predicting the Outcome of rTMS Depression Therapy using EEG Signals and CNN}

\titlerunning{Predicting the Outcome of rTMS Depression Therapy}
\author{Wael Korani  \inst{1} \and 
Md Fahimul Kabir Chowdhury\inst{2} \and 
Sadam AlQadi \inst{3} \and 
Priyan Malarvizhi kumar \inst{1}
\and Reza Rostami \inst{4} \and Reza Kazemi \inst{4}
}
\institute{Department of Data science, University of North Texas, Texas, USA,
\\
\email{Wael.Korani@unt.edu} and \email{Priyan.Malarvizhikumar@unt.edu}\\
\and
Department of Computer Science, University of North Texas, Texas, USA,
\and 
Department of Information science, University of North Texas, Texas, USA,\\
\and
Department of Psychiatry, University of Tehran, Tehran, Iran \\
}

\maketitle              
\begin{abstract}
Repetitive transcranial magnetic stimulation (rTMS) is a non invasive therapy for Major Depressive Disorder (MDD). In this study, we generate images using two time frequency methods to represent EEG signals: Fourier-Bessel Series Expansion with Euclidean Distance (FBSE-ED) and Discrete Wavelet Transform (DWT). We propose an efficient deep learning classifier to predict the outcome of rTMS depression therapy. In this study, we use a private rTMS databases to train a lightweight custom Convolutional Neural Network (CNN) using 10-fold cross validation strategy in order to avoid any bias in our results. The results show that the FBSE-ED representation achieves the highest classification accuracy of 93.60\%, outperforming traditional time-frequency technique (DWT). In addition, the proposed architecture with FBSE-ED image representation technique outperforms more complex EEG-Specific deep learning models (EEGNet, DeepConvNet, SleepEEGNet) by 3.62-10.72\% and pretrained models (Xception, DenseNet201, and MobileNetV2) by 23.03-27.35\%. For more experiments, we utilize another private rTMS database as test database to show the robustness of the proposed model. Our results suggest that integrating advanced signal decomposition with deep learning can facilitate early prediction of rTMS treatment response and support more targeted clinical decision-making. The proposed framework is interpretable, computationally efficient, and well-suited for deployment in real-world local psychiatric clinics.

\end{abstract}
\section{Introduction}

Major Depressive Disorder (MDD) is a complex and debilitating mental health condition that impacts people of all ages and socioeconomic statuses. Its etiology is complex, involving an interplay of genetic predisposition, neurobiological abnormalities, environmental stressors, and psychological influences. Major underlying mechanisms involve imbalances in neurotransmitter activity, disruptions in the functioning of the hypothalamic-pituitary-adrenal (HPA) axis, and changes in the structure of brain areas responsible for regulating emotions. Clinically, MDD manifests through a broad spectrum of symptoms, including persistent low mood, anhedonia, cognitive deficits. In severe cases, suicidal ideation substantially impairing an individual's social, occupational, and personal functioning. While several therapeutic involvements including repetitive transcranial magnetic stimulation (rTMS), psychotherapy, and selective serotonin reuptake inhibitors (SSRIs) are available, treatment outcomes vary significantly across patients, highlighting the need for personalized approaches.

Recent developments in both machine learning (ML) and deep learning (DL) are reshaping approaches, transformative tools in the domain of psychiatric disorders, particularly in the resolution and treatment of MDD. ML techniques are capable of processing large-scale datasets including clinical records, brain imaging, and genomic sources to uncover markers and trends associated with MDD severity and progression. As a specialized branch of ML, DL employs multi-layered artificial neural networks that offer increased capacity to model complex, non-linear relationships. DL has demonstrated strong performance in extracting meaningful features from neurophysiological signals such as EEG and neuroimaging modalities like functional MRI (fMRI), thereby enhancing our understanding of the neurobiological underpinnings of depression and treatment responsiveness. Apart from improving clinical results, ML and DL methodologies can aid in the identification of novel therapeutic routes and encourage the creation of creative therapeutic approaches.

Our central focus in this work is to predict the treatment response status of patients to rTMS therapy classified as responders (R) and non-responders (NR) in the context of MDD. To achieve this, we investigate two distinct EEG-based image representations: (1) time-frequency images generated via Discrete Wavelet Transform (DWT) and (2) FBSE-ED images, derived from Fourier-Bessel Series Expansion combined with Euclidean Distance analysis. Generates time-frequency images using the two methods, yielding two distinct datasets. The datasets are subsequently utilized to train a uniquely structured Convolutional Neural Network (CNN).

\vspace{1em}
\noindent The primary contributions of this work are as follows:
\begin{itemize}
\item We construct two types of EEG-derived image datasets: FBSE-ED images generated using Fourier-Bessel Series Expansion and Euclidean Distance, and time-frequency images created using Discrete Wavelet Transform (DWT).

\item We propose a practical and lightweight custom CNN architecture designed to classify rTMS response outcomes in MDD patients as responders or non-responders.

\item We perform a comparative analysis of the two EEG image representations (FBSE-ED vs. DWT) to identify the most effective modality for predicting rTMS therapy outcomes using our proposed CNN, state-of-the-art pretrained architectures and deep learning models. 

\item We reduce computational overhead by avoiding heavy preprocessing, deep network structures, extensive data augmentation, and reliance on pretrained models.

\item We evaluate model performance using an unbiased cross-validation (CV) strategy, and assess performance using metrics including F1-score, recall, accuracy (acc), and precision to identify the best-performing configuration for predicting depression treatment outcomes.

\end{itemize}

\section{Related Work}
In~\cite{acharya2018automated}, Acharya et al. proposed an creative way for depression detection based on EEG, employing a CNN that autonomously extracts features from raw EEG data without requiring handcrafted input. The model was trained and evaluated on EEG recordings from 30 subjects where 15 with depression and 15 healthy controls. Classification performance using EEG signals from both hemispheres and found that right hemisphere EEG signals yielded higher accuracy, achieving 96.0\%, while the left hemisphere resulted in 93.5\%. In ~\cite{akbari2024prediction} Hesam et al. introduced a method for predicting rTMS medication outcomes in MDD patients using multi-dimensional EEG signals, involving the extraction of both linear and nonlinear attributes for analytical purposes. The researchers tested several ML and neural network (NN) models, ultimately identifying the Cascade Forward Neural Network (CFNN) as the top-performing architecture. Their proposed CFNN model, comprised 11 layers, each containing 10 neurons, and attained a classification accuracy of 97.10\%. Ten-fold cross-validation was put to use to promote generalization and minimize the risk of overfitting. In ~\cite{shovon2019classification} Shovon et al. used EEG data from two well-known sources, BCI Competition IV dataset 2b and BCI Competition II dataset III to classify motor imagery signals. For classification, they proposed a multi-input CNN to improve model generalization. The network was designed with three input streams to process EEG channels independently. The best classification accuracy reached 97.7\% for subject 7 in dataset III, while the average accuracy on dataset 2b was 89.19\%, demonstrating strong performance across datasets.


In ~\cite{mirjebreili2024prediction} Mirjebreili et al. analyzed EEG data from 30 MDD patients, focusing on pre-treatment recordings to distinguish between SSRI responders and non-responders. The researchers first computed effective connectivity across delta, theta, alpha, and beta bands, transforming these into image representations. They employed a hybrid deep learning approach combining BiLSTM with various transfer learning models to utilize both spatial and temporal patterns in the experimental materials. Where BiLSTM achieved the best achievement, reaching an accuracy of 98.33\%. In ~\cite{ebrahimzadeh2021predicting} Elias et al. utilized resting-state EEG recordings from 50 individuals diagnosed with treatment-resistant depression (TRD) and 24 non-depressed healthy controls (HC) to evaluate treatment response following 4-6 weeks of rTMS employed to the dorsolateral prefrontal cortex (DLPFC). The researchers implemented three classification models such as Support Vector Machine (SVM), K-Nearest Neighbor (KNN) and Artificial Neural Network (ANN)  
to distinguish between depression and normal states based on EEG patterns. The SVM model performed best, achieving an accuracy of 82.43\%, followed by ANN (78.37\%) and KNN (74.32\%). In ~\cite{xu2023depressive} Xu et al. used resting-state EEG recordings from 41 depressive disorder (DD) and a control group of 34 healthy subjects, focusing on signals acquired from only six frontal channels to ensure practical and efficient data collection. For classification, the authors developed two distinct deep learning architectures: a Multi-Resolution CNN integrated with LSTM (MRCNN-LSTM), and another incorporating model Residual Squeeze-and-Excitation modules (MRCNN-RSE). The latter model, MRCNN-RSE, yielded superior performance with an accuracy of 98.48\% ± 0.22\%. 


In ~\cite{hasanzadeh2019prediction} Fatemeh et al. utilized resting-state EEG data from 46 MDD patients collected via 19 electrodes before undergoing a 7-week rTMS therapy course. They extracted a wide range of features, including Lempel-Ziv Complexity (LZC), Katz Fractal Dimension (KFD), power spectral density, Correlation Dimension (CD), bispectrum-based features, and cordance measures. Feature relevance was refined using the minimal-redundancy-maximal-relevance (mRMR) technique. For classification, the study implemented a KNN model, evaluated properly through leave-one-out cross-validation. Their method accomplished 91.3\% classification accuracy, a maximum, with equal values for specificity and sensitivity. In ~\cite{ebrahimzadeh2023machine} Elias et al. used resting-state EEG recordings from 88 MDD patients, acquired through a 32-electrode setup prior to a 7-week rTMS therapy. To predict responder status, three models as SVM, KNN, and Multilayer Perceptron (MLP) were trained and evaluated using 10-fold cross-validation. Among them, SVM achieved the best classification accuracy of 94.31\%, along with a sensitivity of 95.65\%, specificity of 92.85\%, and precision of 92.85\%. In ~\cite{zhao2025predicting} Zaho et al. involved a large cohort of 117 MDD patients, comprising 74 responders and 43 non-responders to rTMS therapy. The researchers evaluated eight baseline EEG and clinical elements using seven different state-of-the-art machine learning models, aiming to improve prediction accuracy for treatment response. Among the tested methods, SVM achieved the maximum of 97.33\% accuracy when using a combined feature set of phase locking value (PLV) and clinical variables.

\section{Materials and Methods}

\begin{figure}[ht]
\centering
\includegraphics[width=.93\textwidth]{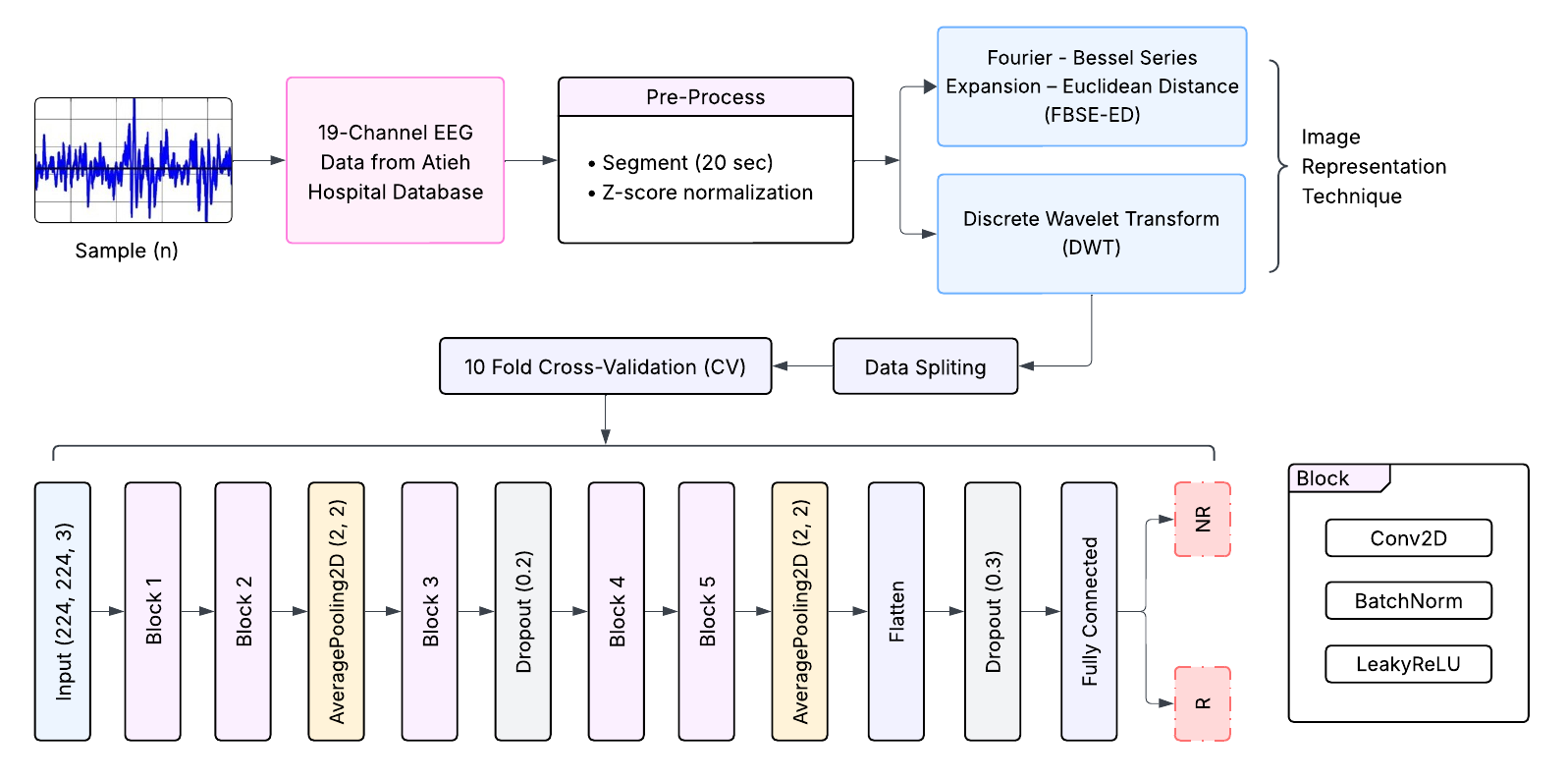}
\caption{The process of pre-processing the EEG signals and generating the images using
different time frequency methods to propose a custom CNN}
\label{blockD}
\end{figure}

Our experiments are carried out on a powerful computer system that has two NVIDIA H100 NVL GPUs, each with 95,830 MB of memory, supported by CUDA 12.2. The machine’s substantial output computational capabilities enabled with efficient training, fine-tuning, and evaluation of all DL models. Figure~\ref{blockD} presents a high-level depiction of the architecture we propose.

\subsection{EEG Data Acquisition}

The rTMS therapy datasets used in this study was collected directly from a private dataset sourced at Atieh Hospital located in Tehran, Iran. Clinical diagnoses of depression were made by psychologist Dr. Reza Kazemi and psychiatrist Dr. Reza Rostami, who also prescribed rTMS treatment to the patients. 

To determine treatment outcomes, the Beck Depression Inventory (BDI) scores were measured for each patient before and four weeks after therapy. Individuals were categorized as responders (R) if their BDI scores decreased by at least 50\%, while those failing to meet this improvement threshold were categorized as non-responders (NR). These classifications were further verified by the attending psychiatrist and psychologist. A summary of BDI scores for all subjects is presented in Table~\ref{table:bdi_scores}.

EEG recordings were captured using the standard 10–20 international electrode placement system, employing 19 scalp electrodes. Signals were sampled at 500 Hz. Figure~\ref{Fig:channel-brain} depicts the electrode configuration used. For signal uniformity across datasets, EEG recordings were segmented into fixed-length windows of 1024 samples (approximately 2.05 seconds). It is noteworthy that the A1–A2 reference channel pair is not available in this dataset.

And later on, we use a much larger dataset to test our model which comprises pre-treatment EEG recordings from 40 individuals. Prior to the initiation of rTMS therapy, each patient’s depressive symptoms were assessed using the BDI. Following a four-week rTMS treatment course, BDI scores were reassessed. Patients exhibiting at least a 50\% reduction in BDI scores were classified as responders, while those below this threshold were considered non-responders. These clinical assessments were further validated by the medical professionals involved. 

\begin{table}[t]
\centering
\caption{BDI score of patients in Atieh Hospital database for rTMS therapy.}
\label{table:bdi_scores}
\begin{tabular*}{\linewidth}{@{\extracolsep{\fill}}lcccc@{}}
\toprule
\textbf{Patient ID} & \textbf{Pre BDI Score} & \textbf{Post BDI Score} & \textbf{Respond to therapy} \\
\midrule
1   & 34 & 11 & R \\
2   & 22 & 0  & R \\
3   & 29 & 8  & R \\
4   & 28 & 12 & R \\
5   & 36 & 7  & R \\
6   & 11 & 3  & R \\
7   & 40 & 16 & R \\
8   & 22 & 8  & R \\
9   & 12 & 2  & R \\
10  & 26 & 21 & NR \\
11  & 29 & 16 & NR \\
12  & 16 & 11 & NR \\
13  & 48 & 29 & NR \\
14  & 24 & 13 & NR \\
15  & 56 & 45 & NR \\
\bottomrule
\end{tabular*}
\end{table}


\begin{figure}[t]
\centering

\begin{minipage}{0.48\textwidth}
    \centering
    \includegraphics[width=\linewidth]{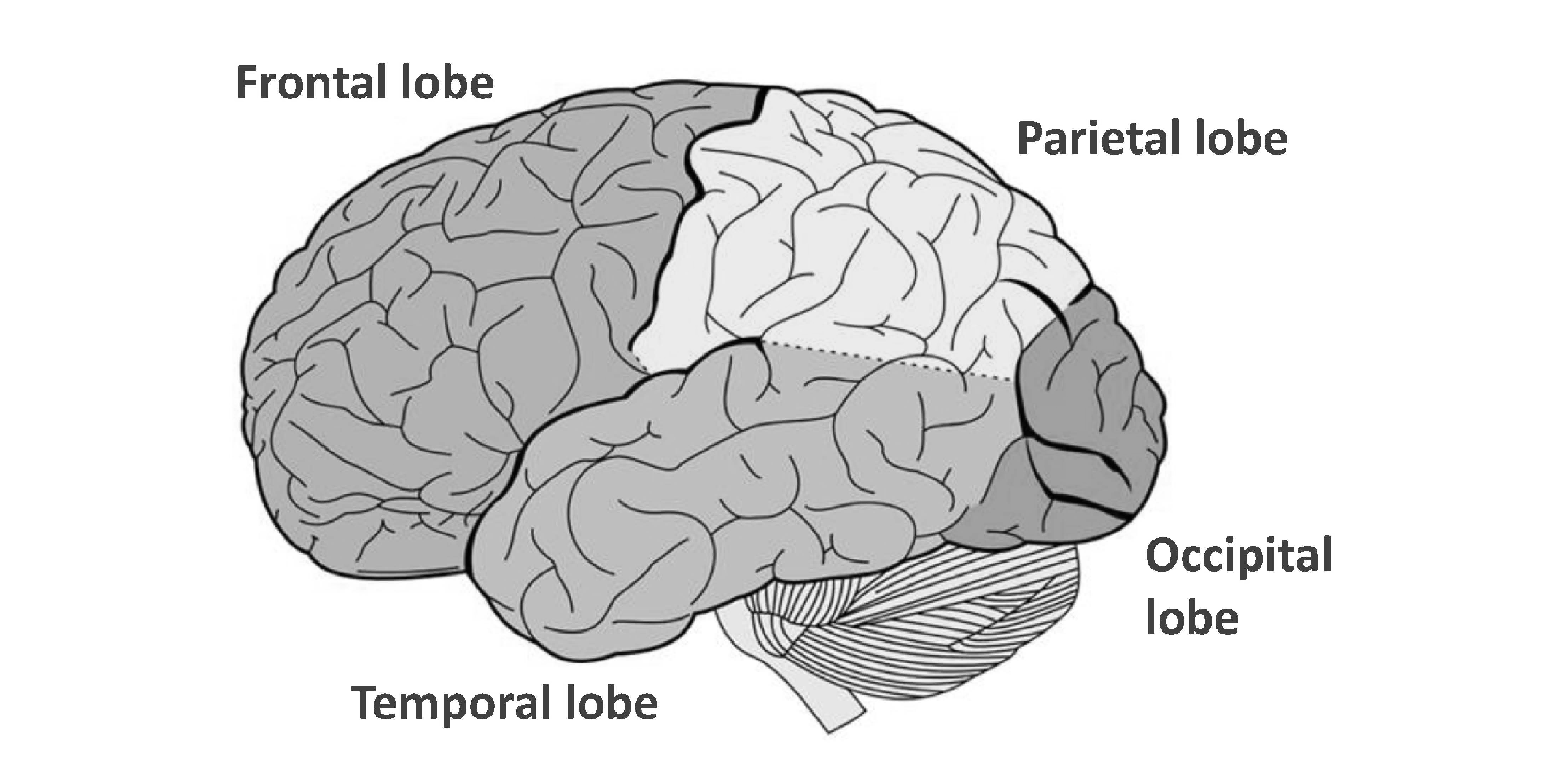}
    \caption*{(a)}
\end{minipage}
\hfill
\begin{minipage}{0.48\textwidth}
    \centering
    \includegraphics[width=\linewidth]{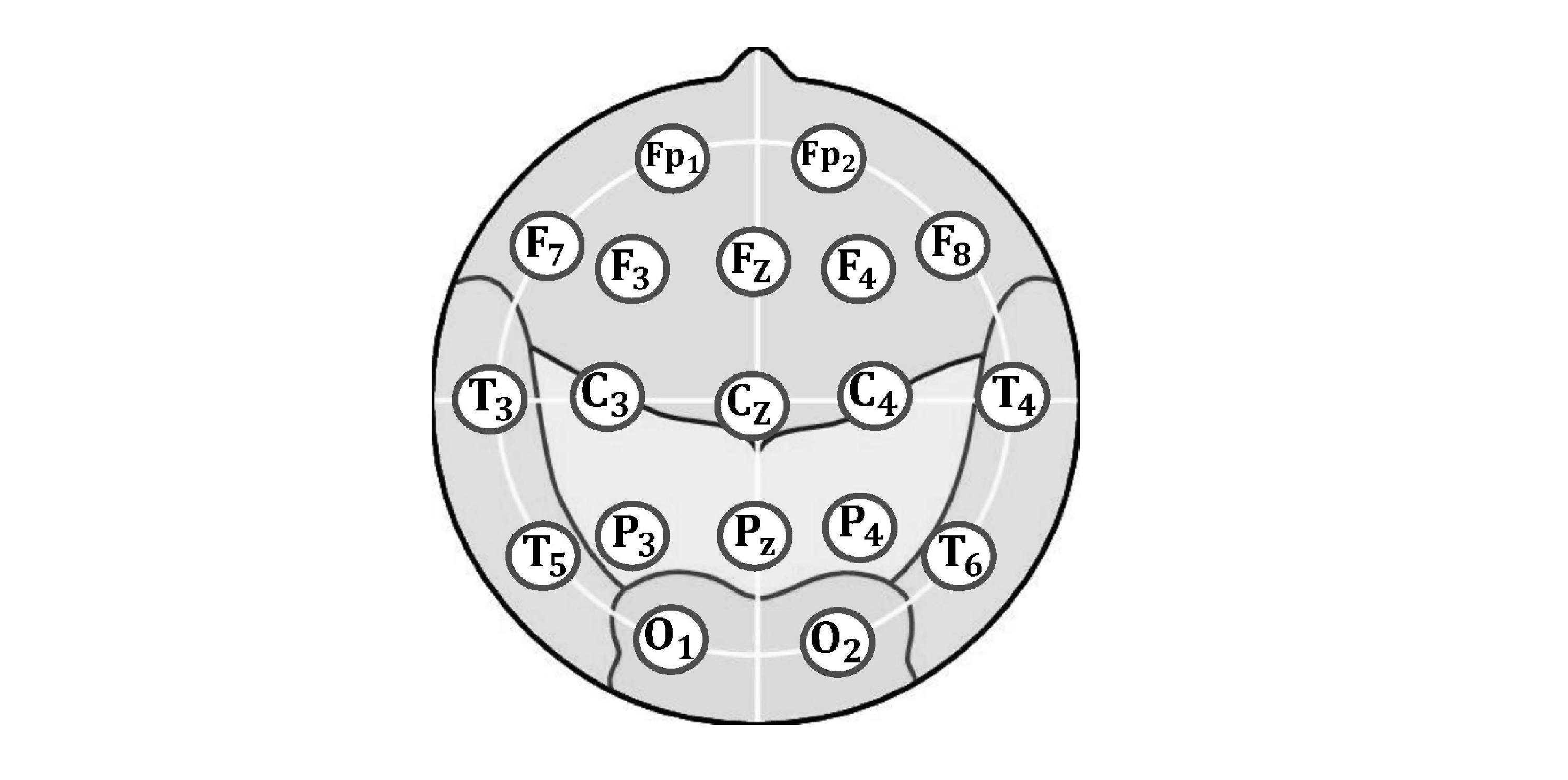}
    \caption*{(b)}
\end{minipage}

\caption{Channel positions according to their mapping onto brain lobes (left) and the 10--20 EEG system (right).}
\label{fig:channel-brain}
\end{figure}

\subsection{Time-frequency methods}

Convolutional Neural Networks (CNNs) are employed to classify EEG signals by first converting them into visual formats using two separate time-frequency transformation techniques. These methods generate time-frequency images from raw EEG data, which are subsequently used as inputs to train the CNN model. The models demonstrated effective training and improved learning dynamics when incorporating these representations, leading to enhanced classification performance. In this study, we specifically employ two time-frequency analysis methods: Discrete Wavelet Transform (DWT) and Fourier-Bessel Series Expansion with Euclidean Distance (FBSE-ED).

\subsubsection{Fourier-Bessel Series Expansion (FBSE)}

The Fourier-Bessel Series Expansion (FBSE) is an effective method for representing non-stationary signals due to its inherently non-stationary basis functions. Compared to the traditional Fourier transform, FBSE provides enhanced spectral resolution. The synthesis of a signal \( y(n) \), where \( n = 0, 1, \dots, V-1 \), using the zero-order FBSE can be written as:

\begin{equation}
y(n) = \sum_{k=1}^{V} D_k J_0\left(\frac{\lambda_k n}{V}\right)
\end{equation}

Here, \( D_k \) are the FBSE coefficients, calculated using:

\begin{equation}
D_k = \frac{2}{V^2 [J_1(\lambda_k)]^2} \sum_{n=0}^{V-1} n y(n) J_0\left(\frac{\lambda_k n}{V}\right)
\end{equation}

In these expressions, \( J_0(\cdot) \) and \( J_1(\cdot) \) represent the zeroth- and first-order Bessel functions, respectively, and \( \lambda_k \) denotes the \( k \)-th positive root of \( J_0 \). These roots are typically computed using numerical techniques such as the Newton-Raphson method. The link between the FBSE index \( k \) and the corresponding continuous-time frequency \( f_k \) (in Hz) is given by:

\begin{equation}
k = \frac{2 f_k V}{f_s}
\end{equation}

where \( f_s \) is the signal's sampling frequency. To fully span the signal's bandwidth, \( V \) is chosen to equal the signal length, which is set to 4097 in this study.

\subsubsection{Rhythms Separation}

EEG signals are typically broken down into five distinct and selected frequency bands: alpha (8–13 Hz), beta (13–30 Hz), gamma (30–86.81 Hz), theta (4–8 Hz), and delta (0.5–4 Hz). Within the FBSE framework, these rhythms can be isolated by identifying the appropriate index intervals, derived from the relation between FBSE root order and frequency. The coefficient index ranges for each rhythm are defined as follows: [24, 188], [189, 377], [378, 613], [614, 1416], [1417, 4097].

The signal can then be decomposed as:

\begin{align}
y(n) &= \sum_{k=\delta_L}^{\delta_U} D_k J_0\left(\frac{\lambda_k n}{V}\right) +
       \sum_{k=\theta_L}^{\theta_U} D_k J_0\left(\frac{\lambda_k n}{V}\right) +
       \sum_{k=\alpha_L}^{\alpha_U} D_k J_0\left(\frac{\lambda_k n}{V}\right) \nonumber \\
     &+ \sum_{k=\beta_L}^{\beta_U} D_k J_0\left(\frac{\lambda_k n}{V}\right) +
       \sum_{k=\gamma_L}^{\gamma_U} D_k J_0\left(\frac{\lambda_k n}{V}\right)
\end{align}

Each summation corresponds to a specific EEG rhythm. Coefficients outside these defined ranges are set to zero to reconstruct the individual rhythm components in the time domain. The coefficient sets CS1 through CS5 correspond to the delta through gamma rhythms, respectively.

A notable advantage of FBSE-based rhythm decomposition is its single-step process, in contrast to the wavelet-based approach which typically requires multi-level decomposition. This makes FBSE rhythm separation more efficient and easier to implement.

\subsubsection{Euclidean Distance}

Consider two points having \( A = (a_1, a_2, \dots, a_n) \) and \( B = (b_1, b_2, \dots, b_n) \) located in an \( n \)-dimensional Euclidean space. The straight-line distance, also known as the Euclidean distance, between these two points is calculated using the following equation~\cite{dattorro2010convex}:

\begin{equation}
\text{Euclidean Distance} = \sqrt{\sum_{i=1}^{n}(b_i - a_i)^2}
\end{equation}

In the context of our current study, we consider a 2-dimensional space, hence \( n = 2 \). For simplification, we define point \( B \) as the origin, i.e., \( B = (0, 0) \), which is a common reference point in Euclidean geometry~\cite{eisenhart2005coordinate}.

This distance measure is applied to each rhythm component obtained after performing rhythm separation using FBSE. Once the Euclidean distance values are computed for the EEG rhythms corresponding to seizure and normal classes, the resulting matrices are visualized as grayscale images. These representations, shown in Fig.~\ref{EDU_images}, reveal distinguishable patterns between the two classes, thereby supporting the discriminative utility of Euclidean distance in our rhythm-based EEG analysis framework.

\subsubsection{Discrete Wavelet Transform (DWT)}
The DWT is a strong technique used for time-frequency analysis, allowing a signal to be decomposed into components at different frequency bands and temporal resolutions. In contrast to the Continuous Wavelet Transform (CWT), which analyzes signals across a continuous range of scales, the Discrete Wavelet Transform (DWT) operates at distinct scale intervals. It achieves this by recursively applying pairs of filters with high-pass and low-pass to the signal, resulting in a layered decomposition that captures both frequency and temporal details at multiple resolutions. This framework enables effective analysis of localized events and sudden changes within the signal. While the DWT may exhibit reduced time resolution at higher frequencies, its efficiency and effectiveness in isolating essential signal features make it a valuable tool in areas such as denoising, compression, and feature extraction.

\subsection{EEG signal preprocessing}

\begin{figure}[t]
\centering

\begin{minipage}[b]{0.48\linewidth}
    \centering
    \includegraphics[width=\linewidth]{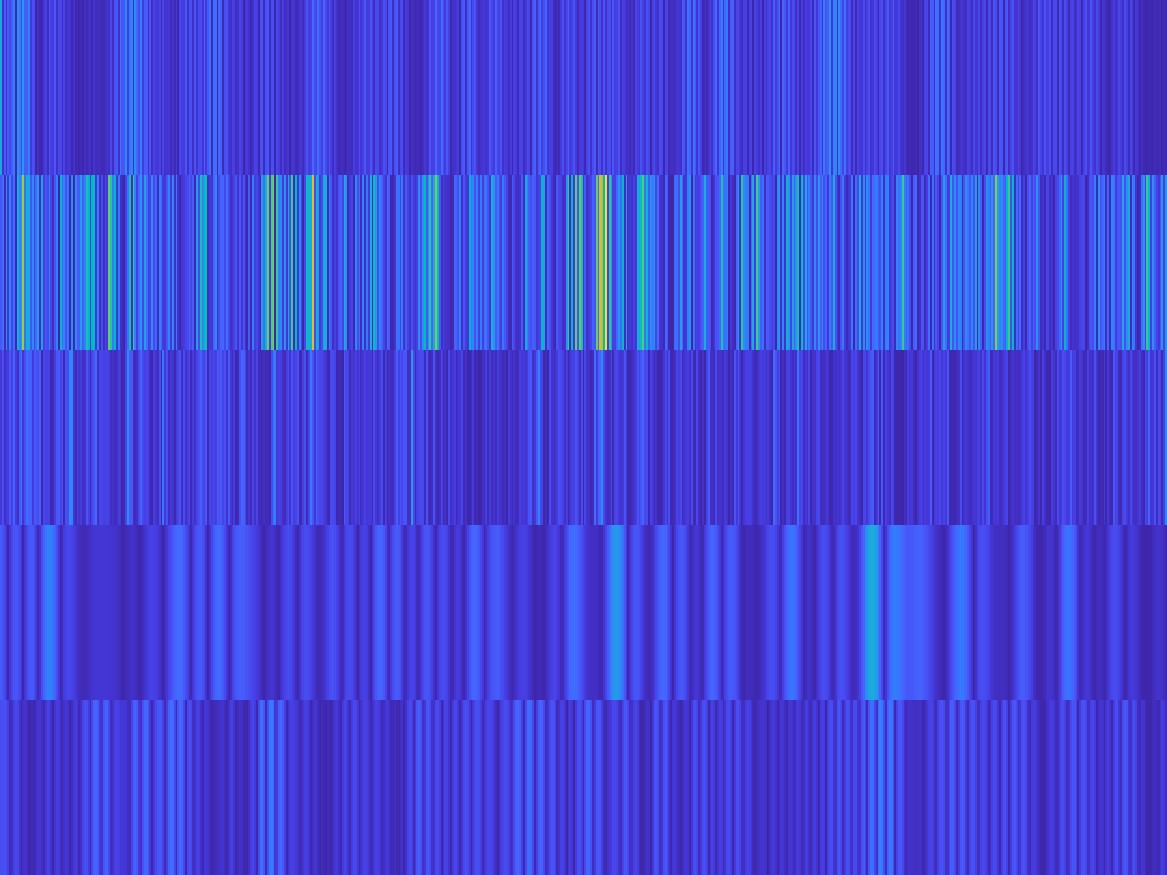}
    
    \small (a) Responder
\end{minipage}
\hfill
\begin{minipage}[b]{0.48\linewidth}
    \centering
    \includegraphics[width=\linewidth]{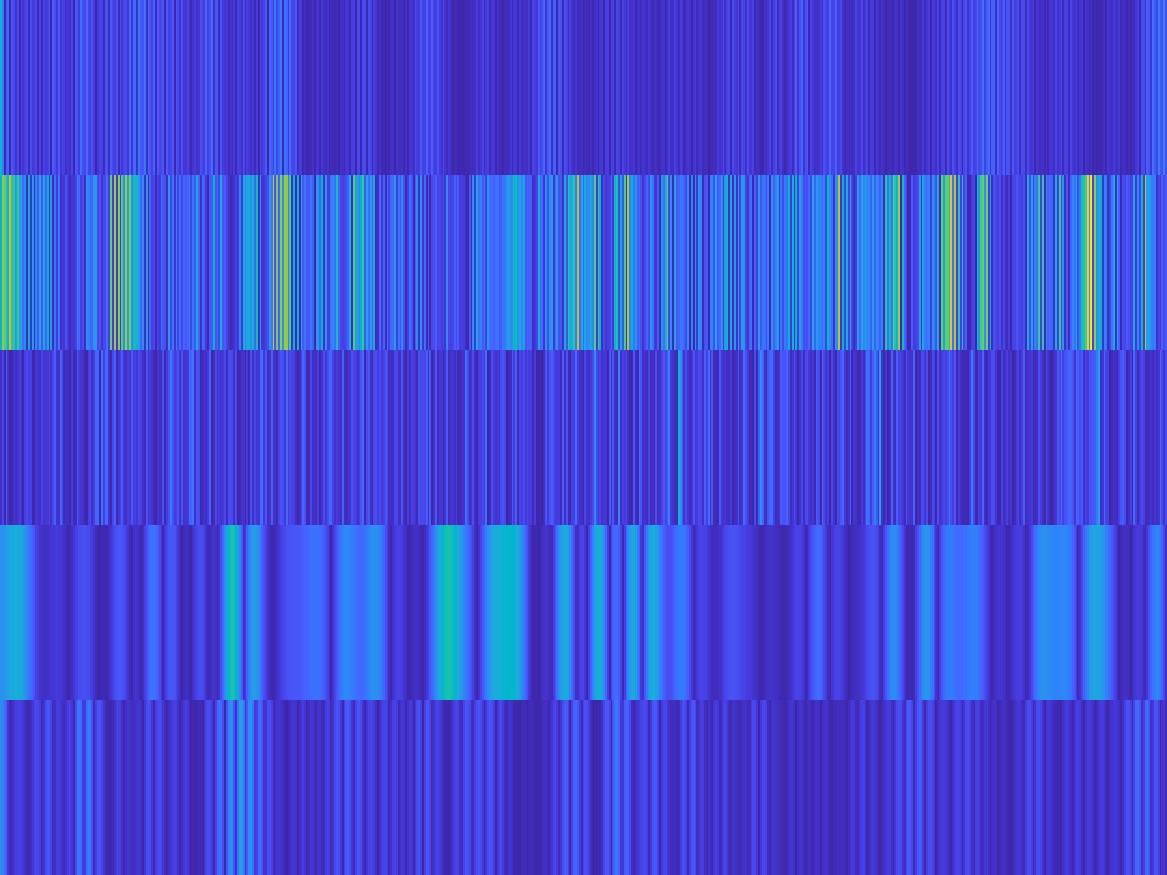}
    
    \small (b) Non-Responder
\end{minipage}

\caption{Sample FBSE-ED-based representations of a responder and a non-responder.}
\label{fig:edu_images}
\end{figure}

To generate FBSE-ED images, every EEG channel is first segmented into 20-second intervals following standard preprocessing steps, which include band-pass filtering, notch filtering. The Fourier-Bessel Series Expansion (FBSE) is then applied to each segment to decompose the signal into distinct frequency rhythms. For each rhythm, spatial representations are derived by calculating the Euclidean distance from the origin, producing matrices with non-negative values. These matrices are subsequently converted into grayscale images, capturing the spatial features of brain rhythms. The collection of these FBSE-ED images constitutes a dataset that is used to train models aimed at predicting rTMS treatment response in patients with MDD. Example FBSE-ED images for both groupings of responders and non-respondents are present in Fig.~\ref{fig:edu_images}.

\begin{figure}[t]
\centering

\begin{minipage}[b]{0.48\linewidth}
    \centering
    \includegraphics[width=\linewidth]{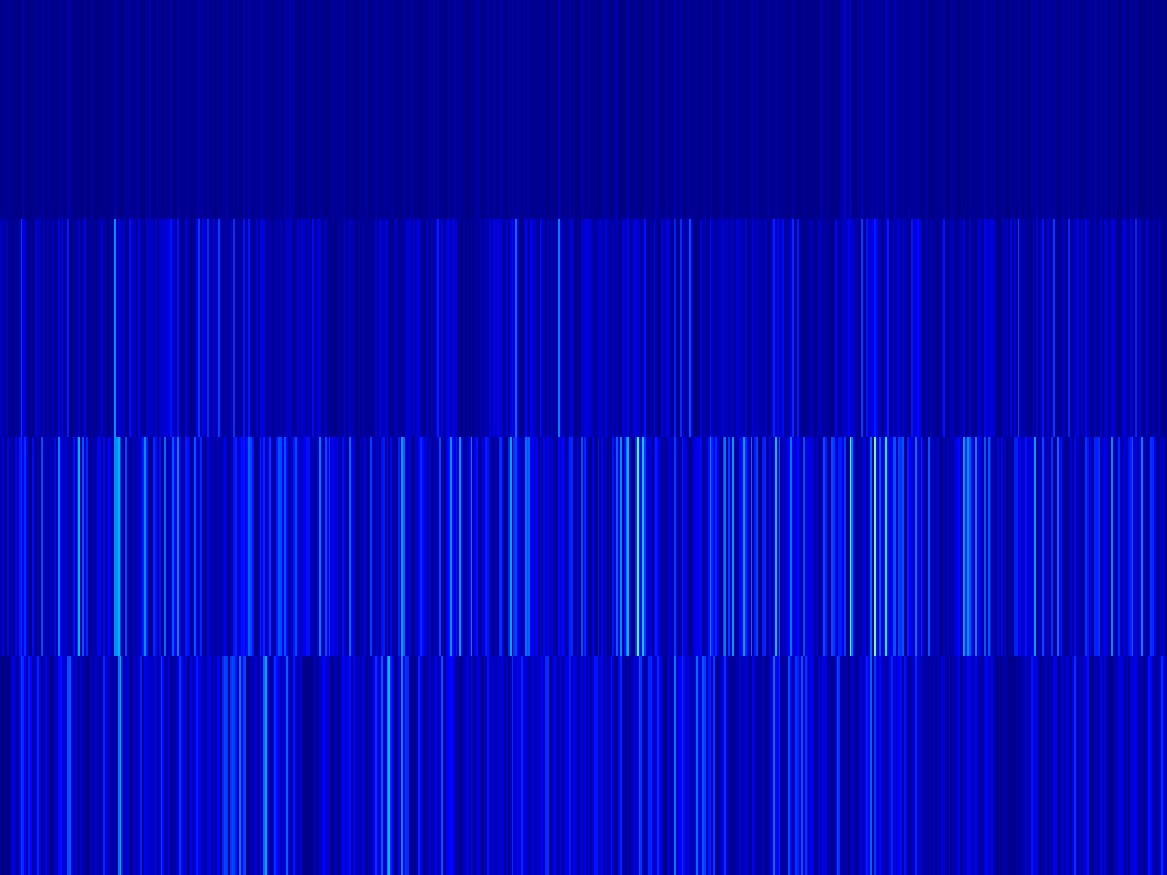}
    \small (a) Responder
\end{minipage}
\hfill
\begin{minipage}[b]{0.48\linewidth}
    \centering
    \includegraphics[width=\linewidth]{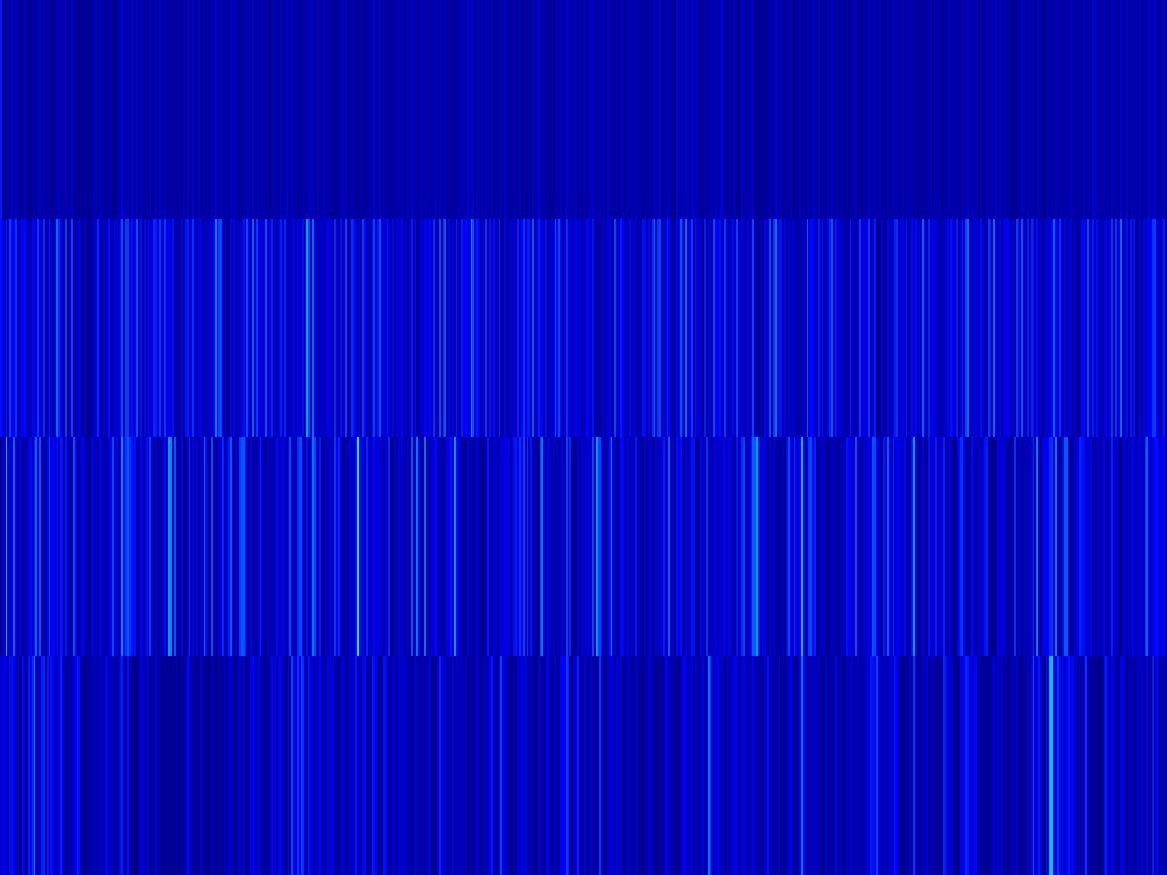}
    \small (b) Non-Responder
\end{minipage}

\caption{Sample DWT images of a responder and a non-responder.}
\label{fig:dwt_images}
\end{figure}

To generate time-frequency representations using the Discrete Wavelet Transform (DWT), EEG signals were first segmented into 20-second epochs. Each segment was then processed using DWT with the ‘db4’ wavelet and a decomposition level of 5. This multilevel breakdown excerpts signal elements across varying frequency bands, enabling the construction of detailed time-frequency images that capture both lower and higher frequency components activity. The DWT was applied following standard preprocessing steps, involved filtering signals between 0.1 and 70 Hz, removing 50 Hz line noise using a notch filter, and resampling the data at 128 Hz to lessen processing demands. DWT enables multi-scale frequency analysis of EEG signals, which is advantageous for detecting unique signal patterns associated with response and non-response to treatment. The resulting DWT-based time-frequency images form the basis of our dataset, with representative samples illustrated in Fig.~\ref{fig:dwt_images}.


\section{Proposed CNN}

The CNN architecture proposed in this study is designed for binary classification and incorporates a deep convolutional structure optimized for effective feature extraction from time-frequency EEG images. The model begins with an input layer set up for 224x224 images with three channels. In the first layer, 32 convolutional filters of size 5×5 are applied with a (2,2) stride and 'same' padding, followed by batch normalization and a LeakyReLU activation. This pattern is repeated with 64 filters in the second convolutional layer, stride (1,1), and an additional average pooling layer with a window of 2×2. Subsequent layers further deepen the network: the third convolutional layer introduces 128 filters with a windoq of (2,2) stride, followed by batch normalization, using LeakyReLU, and a dropout layer to reduce overfitting at a rate of 0.2. The fourth convolutional layer adds 256 filters, and the fifth applies 512 filters, both using 5×5 kernels with stride settings of (2,2) and (1,1) respectively, each followed by batch normalization and LeakyReLU. Another average pooling layer is included after the final convolutional block.

After the convolutional feature extraction stages, the output is flattened and passed through a 0.3 rate of dropout layer. A dense layer with a sigmoid activation function handles binary classification. The model is optimized using Adam (learning rate: 0.0001), where the loss function is binary cross-entropy. To ensure reliable and unbiased performance evaluation, we employ a 10-fold cross-validation strategy. We then randomly partitioned the entire dataset into 10 mutually exclusive folds using the KFold method with shuffling and a fixed random seed is used of 42 to ensure reproducibility. For each fold in 10-fold cross-validation, the process is performed ten times, with the model being trained on nine folds and tested on the last one. Performance is evaluated using accuracy (acc), and training is monitored with early stopping to prevent overfitting. A specific summary of this framework is shown in Fig~\ref{blockD}, and hyperparameters are listed in Table~\ref{Training_Hyperparameters}.

\begin{table}[t]
\centering
\caption{Training Hyperparameters Across Model Architectures}
\label{Training_Hyperparameters}
\begin{tabular*}{\linewidth}{@{\extracolsep{\fill}}lccc@{}}
\toprule
\textbf{Hyperparameters} & \textbf{EEG Models} & \textbf{Pretrained} & \textbf{Proposed CNN} \\
\midrule
Input shape & (224, 224, 3) & (224, 224, 3) & (224, 224, 3) \\
Batch size & 32 & 32 & 32 \\
Dropout rate & 0.2, 0.3 & 0.5, 0.3 & 0.2, 0.3 \\
Learning rate & 0.0001 & 0.0001 & 0.0001 \\
Optimizer & Adam & Adam & Adam \\
Activation & LeakyReLU & LeakyReLU & LeakyReLU \\
Cross-Validation & 10-fold & 10-fold & 10-fold \\
\bottomrule
\end{tabular*}
\end{table}

\subsection{Transfer Learning (TL) Architecture}

We employ a transfer learning strategy using ImageNet-pretrained weights for effective feature extraction. The top classification layers are removed, and the base model is frozen to retain learned representations. Spatial features are aggregated using a Global Average Pooling (GAP) layer, improving generalization. It is followed by Batch Normalization and to prevent overfitting a Dropout layer (rate = 0.5) has been used. A Dense layer with 256 units and LeakyReLU activation captures complex patterns, followed by another Dropout (rate = 0.3) for regularization. The final sigmoid-activated neuron performs binary classification between responder (R) and non-responder (NR). The TL models are consistently trained at a learning rate of 0.0001 using the Adam optimizer and evaluated using 10-fold cross-validation. Early stopping based on the achieved validation loss is applied to cut of the overfitting. A detailed summary of the hyperparameters and training configuration is presented in Table~\ref{Training_Hyperparameters}.

\subsection{EEG-Specific Baseline Models}
To benchmark our proposed approach against established EEG-based models, we implement and evaluate three widely used deep learning architectures: EEGNet, DeepConvNet, and SleepEEGNet. EEGNet~\cite{lawhern2018eegnet} employs compact depthwise and separable convolutions to efficiently extract temporal and spatial EEG features, making it suitable for low-resource settings. DeepConvNet~\cite{schirrmeister2017deep}, on the other hand, utilizes a deeper stack of convolutional layers with max-pooling and dropout for robust feature learning across EEG channels. SleepEEGNet~\cite{mousavi2019sleepeegnet} is a variant tailored for sleep stage classification, combining temporal convolutions and hierarchical feature extraction. All three models were adapted to accept time-frequency transformed EEG inputs of shape (224×224×3) and were trained under identical cross-validation settings for consistent comparison.

\subsection{Performance Evaluation}

Although numerous metrics exist to evaluate the performance of classification models, accuracy remains a core and commonly adopted criterion. In this work, we focus primarily on accuracy as the principal evaluation metric. The percentage of correctly predicted cases compared to the entire amount of samples assessed is known as accuracy, and it provides a gauge of the model's overall dependability and predictive capacity.

\begin{equation}
\text{Accuracy} = \frac{TP + TN}{TP + FP + TN + FN}
\end{equation}

where $TP$ is indicated as true positives, $TN$ is indicated true negatives, $FP$ is false positives, false negative is indicated as $FN$.

Given the imbalanced nature of the MDD dataset, relying solely on accuracy is insufficient. To gain a more comprehensive evaluation, we utilize the confusion matrix, which provides a tabulated summary of correct and incorrect predictions across both classes. From this, we derive additional key metrics:

\begin{itemize}
    \item The percentage of true positive predictions from all positive forecasts is measured by precision:
    \begin{equation}
    \text{Precision} = \frac{TP}{TP + FP}
    \end{equation}

    \item The model's recall (also known as sensitivity) estimates how well it can detect every real positive case:
    \begin{equation}
    \text{Recall} = \frac{TP}{TP + FN}
    \end{equation}

\end{itemize}

To provide a balanced view of precision and recall, we compute the F-score, which is the harmonic mean of the two. For multi-class or detailed binary analysis, the macro-averaged F-score is calculated as:

\begin{equation}
\text{F-score} = \frac{2}{3} \sum_{c=1}^{3} \frac{\text{Precision}_c \cdot \text{Recall}_c}{\text{Precision}_c + \text{Recall}_c}
\end{equation}

Additionally, we incorporate the Receiver Operating Characteristic (ROC) curve as a visual and quantitative tool to evaluate model performance. The Area Under the Curve (AUC) serves as a critical measure of class separability. A perfect model yields an AUC of 1.0, indicating flawless discrimination between responder and non-responder classes.

\section{Results}

Our proposed CNN model is trained and tested on EEG datasets from MDD patients, which have been preprocessed and converted into time-frequency representations using two distinct transformation techniques: Discrete Wavelet Transform (DWT) and Fourier-Bessel Series Expansion and Euclidean Distance (FBSE-ED). Accuracy, precision, recall, and F1-score are among the evaluation measures used to evaluate the performance of our suggested based model, providing a comprehensive analysis of its classification capabilities across the different time-frequency representations.

In our first experiment, we compare DWT and FBSE-ED time-frequency method using our custom CNN to identify which time-frequency method is more responsive towards treating MDD. FBSE-ED time-frequency method demonstrated superior performance across all metrics. In Table~\ref{tab:dwt vs ed}, DWT showed strong results with 92\% (NR) and 93\% (R) precision and 91.60\% AUC, FBSE-ED achieved marginally better performance with 93\% (NR) and 94\% (R) precision, along with higher recall (90\% NR, 96\% R) and the best accuracy (93.60\%) in this comparison. 

\begin{table}[t]
\centering
\caption{Comparison with DWT and FBSE-ED time-frequecy method}
\label{tab:dwt vs ed}
\begin{tabular}{lcccccccc}
\toprule
\textbf{Method} & \multicolumn{2}{c}{\textbf{Precision (\%)}} & \multicolumn{2}{c}{\textbf{Recall (\%)}} & \multicolumn{2}{c}{\textbf{F1-score (\%)}} & \textbf{AUC} & \textbf{Accuracy} \\
 & NR & R & NR & R & NR & R &  &  \\
\midrule
DWT & 92 & 93 & 88 & 96 & 90 & 94 & 91.60 & 92.56 \\
FBSE-ED & \textbf{93} & \textbf{94} & \textbf{90} & \textbf{96} & \textbf{91} & \textbf{95} & \textbf{92.99} & \textbf{93.60} \\
\bottomrule
\end{tabular}
\end{table}

To verify our custom cnn's feature extraction capability, in our second experiment, we compare our Proposed CNN with various additional cutting-edge deep learning and pretrained models. 

Table~\ref{tab:pretrained_comparison} presents a detailed comparative evaluation of the suggested custom CNN model against well-established several deep learning frameworks, including Xception, MobileNetV2, DenseNet201, EEGNet, SleepEEGNet, and DeepConvNet. Traditional image-based models such as Xception and MobileNetV2 yielded relatively lower performance, achieving accuracies of 66.25\% and 68.02\%, respectively, with AUC scores below 67\%. In contrast, EEG-specific architectures like EEGNet and SleepEEGNet demonstrated substantial improvements, reaching accuracies of 82.88\% and 86.82\%, respectively, and corresponding AUCs of 81.18 and 86.01. Among the baseline models, DeepConvNet shows the best performance, attaining an accuracy of 89.98\% and an AUC of 88.77. Notably, the proposed CNN model outperformed all baselines across all evaluation metrics, achieving the highest accuracy of 93.60\% and the highest AUC of 92.99. It also maintained superior class-wise performance for both non-response (NR) and response (R) categories, with F1-scores of 91\% and 95\%, respectively. These outcomes show how successful the suggested strategy is in accurately distinguishing between NR and R classes, highlighting its robustness and suitability for EEG-based classification tasks.

\begin{table}[t]
\centering
\caption{Comparison of Proposed CNN with Pretrained and EEG-Specific Models}
\label{tab:pretrained_comparison}
\begin{tabular}{lcccccccc}
\toprule
\textbf{Method} & \multicolumn{2}{c}{\textbf{Precision (\%)}} & \multicolumn{2}{c}{\textbf{Recall (\%)}} & \multicolumn{2}{c}{\textbf{F1-score (\%)}} & \textbf{AUC} & \textbf{Acc (\%)} \\
 & NR & R & NR & R & NR & R &  &  \\
\midrule
Xception      & 63 & 68 & 52 & 77 & 57 & 72 & 64.41 & 66.25 \\
MobileNetV2   & 66 & 69 & 52 & 80 & 58 & 74 & 66.01 & 68.02 \\
DenseNet201   & 70 & 71 & 54 & 83 & 61 & 76 & 68.49 & 70.57 \\
EEGNet        & 80 & 85 & 74 & 88 & 77 & 86 & 81.18 & 82.88 \\
SleepEEGNet    & 83 & 89 & 83 & 89 & 83 & 89 & 86.01 & 86.82 \\
DeepConvNet   & 89 & 90 & 84 & 94 & 86 & 92 & 88.77 & 89.98 \\
Proposed CNN  & \textbf{93} & \textbf{94} & \textbf{90} & \textbf{96} & \textbf{91} & \textbf{95} & \textbf{92.99} & \textbf{93.60} \\
\bottomrule
\end{tabular}
\end{table}

In our third experiment, we use another large dataset of FBSE-ED time-frequency method to test our model's performance and we named it as test dataset. Between the primary and test dataset, the primary dataset maintained a slight but consistent advantage over test dataset. While both dataset performed exceptionally well, primary dataset achieved higher accuracy (93.60\% vs 92.13\%). The 1.47\% higher accuracy and 1.25\% better AUC (92.99\% vs 91.74\%) of primary dataset suggest it as the optimal configuration of our architecture shown in Table~\ref{tab:small vs big}.

\begin{table}[t]
\centering
\caption{Comparison of FBSE-ED's Small and Large Dataset}
\label{tab:small vs big}
\begin{tabular}{lcccccccc}
\toprule
\textbf{Dataset} & \multicolumn{2}{c}{\textbf{Precision (\%)}} & \multicolumn{2}{c}{\textbf{Recall (\%)}} & \multicolumn{2}{c}{\textbf{F1-score (\%)}} & \textbf{AUC} & \textbf{Accuracy (\%)} \\
 & NR & R & NR & R & NR & R &  &  \\
\midrule
Test & 92 & 92 & 89 & 94 & 91 & 93 & 91.74 & 92.13 \\
Primary & \textbf{93} & \textbf{94} & \textbf{90} & \textbf{96} & \textbf{91} & \textbf{95} & \textbf{92.99} & \textbf{93.60} \\
\bottomrule
\end{tabular}
\end{table}

Our custom CNN approach, particularly the primary dataset, emerged as the clear winner across all experiments. It demonstrated: (1) 1.04\% better accuracy than DWT, (2) 23.03\% - 3.62\% higher accuracy than the best pretrained and deep learning models, and (3) 1.47\% superior accuracy to its test dataset. These results validate that our custom CNN architecture is more effective than both traditional methods and existing pretrained models for feature extraction task on FBSE-ED time-frequency method.

\section{Conclusion}
\label{conclusion}

In this study, to predict the rTMS depression treatment, we introduce an efficient deep learning framework for individuals diagnosed with MDD. EEG signals are transformed into FBSE-ED and DWT-based time-frequency images, which are then used to assess the discriminative power of these representations through a custom lightweight CNN. The custom CNN model comprises several convolutional layers, along with batch normalization and dropout components, followed by a final output layer, enabling efficient extraction of meaningful features from the input image data. Among the two representations, the FBSE-ED-based approach achieve superior performance, attaining an accuracy of 93.60\%, and outperformed both conventional time-frequency methods and several well defined state-of-the-art pretrained and deep learning models. These findings confirm that the value of integrating signal transformation techniques with deep learning for early prediction and treatment stratification in psychiatric care. The proposed framework is computationally efficient, scalable, and interpretable, potential to be adapted for other neurological and psychiatric conditions that exhibit EEG biomarkers. For future work, we aim to validate our framework on larger, multimodal and multicenter datasets to enhance population-wide generalizability. Additional directions include exploring alternative Euclidean distance measures across EEG rhythms, and the integration of multimodal inputs such as fMRI, clinical records, and genetic profiles.


\bibliographystyle{unsrt}
\bibliography{ref}

\end{document}